\documentclass[a4paper,fleqn]{cas-dc}

\usepackage[authoryear]{natbib}
\usepackage{graphicx}

\begin{document}
\let\WriteBookmarks\relax
\def\floatpagepagefraction{1}
\def\textpagefraction{.001}

\title[mode = title]{LRNet: Change detection of high-resolution remote sensing imagery via strategy of localization-then-refinement}


\author[1]{Huan Zhong}
\ead{zhonghuan@whu.edu.cn}
\address[1]{State Key Laboratory of Information Engineering in Surveying, Mapping and Remote Sensing, Wuhan University, Wuhan 430079, China}

\author[1]{Chen Wu}
\ead{chen.wu@whu.edu.cn}
\cormark[1] 
\cortext[1]{Corresponding author} 

\author[2]{Ziqi Xiao}
\ead{ziqixiao@whu.edu.cn}
\address[2]{School of Computer Science, Wuhan University, Wuhan 430079, China}

\begin{abstract}[S U M M A R Y]
Change detection, as a research hotspot in the field of remote sensing, has witnessed continuous development and progress. However, the discrimination of boundary details remains a significant bottleneck due to the complexity of surrounding elements between change areas and backgrounds. Discriminating the boundaries of large change areas results in misalignment, while connecting boundaries occurs for small change targets. To address the above issues, a novel network based on the localization-then-refinement strategy is proposed in this paper, namely LRNet. LRNet consists of two stages: localization and refinement. In the localization stage, a three-branch encoder simultaneously extracts original image features and their differential features for interactive localization of the position of each change area. To minimize information loss during feature extraction, learnable optimal pooling (LOP) is proposed to replace the widely used max-pooling. Additionally, this process is trainable and contributes to the overall optimization of the network. To effectively interact features from different branches and accurately locate change areas of various sizes, change alignment attention (C2A) and hierarchical change alignment module (HCA) are proposed. In the refinement stage, the localization results from the localization stage are corrected by constraining the change areas and change edges through the edge-area alignment module (E2A). Subsequently, the decoder, combined with the difference features strengthened by C2A in the localization phase, refines change areas of different sizes, ultimately achieving accurate boundary discrimination of change areas. The proposed LRNet outperforms 13 other state-of-the-art methods in terms of comprehensive evaluation metrics and provides the most precise boundary discrimination results on the LEVIR-CD and WHU-CD datasets. The source code of the proposed LRNet is available at \href{https://github.com/Pl-2000/LRNet}{https://github.com/Pl-2000/LRNet}.
\end{abstract}



\maketitle

\section{Introduction}
Remote sensing (RS) image change detection (CD) is a technique aimed at analyzing whether and what kind of changes exist between images acquired from the same area at different times\citep{ref1,ref2}. As a research hotspot in the field of RS, change detection has experienced continuous development and progress. It is widely applied in various practical scenarios, such as post-disaster assessment\citep{ref3,ref4}, environmental monitoring\citep{ref5} and land resource management\citep{ref6,ref7}. With the ongoing evolution of RS technology, the data sources for change detection have become increasingly diverse\citep{ref8,ref9}. High-resolution RS images can provide rich surface detail\citep{ref10}, facilitating fine-grained change detection, and gradually become the primary data source for change detection.

\par Early change detection is primarily achieved by extracting, analyzing, and comparing spectral or spatial information of the same pixel in bi-temporal images, which roughly includes algebra-based methods, image transformation methods, and post-classification methods. Algebra-based methods calculate the spectral differences of the same pixel in bi-temporal images through algebraic operations to obtain a change intensity map, followed by threshold segmentation algorithms to acquire change results. The common ones are image difference (ID)\citep{ref11}, image ratio (IR)\citep{ref12}, and change vector analysis (CVA)\citep{ref13}. These methods only extract shallow spectral information, making it challenging to analyze the complex spectral features. Image transformation methods analyze and compare the bi-temporal images by transforming them to the identical higher-dimensional feature space, such as principal component analysis (PCA)\citep{ref14,ref15}, multivariate alteration detection (MAD)\citep{ref16}, and slow feature analysis (SFA)\citep{ref17,ref18}. This category of methods utilizes deep spectral information for change detection and improves the detection accuracy. Post-classification methods classify land cover in different temporal images separately and then compare the differences to detect changes. Examples include support vector machines (SVM)\citep{ref19} and random forests (RF)\citep{ref20}. These methods can mitigate changes due to varying environmental conditions during imaging, but the detection performance is significantly influenced by classification accuracy. The aforementioned early change detection methods often require manually designed feature extractors, making it challenging to exploit deep, high-dimensional features in images and greatly limiting the development of change detection.

\par In recent years, deep learning (DL) has experienced flourishing development. It has achieved great success in multiple research fields of computer vision, including classification\citep{ref21,ref22}, object detection\citep{ref23,ref24}, semantic segmentation\citep{ref25,ref26}, medical diagnosis\citep{ref_medical}, and so on. Its excellent feature representation capability and high level of automation have attracted the attention of researchers in the field of CD. A series of DL-based change detection methods have been proposed to improve the performance of change detection. These methods can be roughly categorized into three types in terms of feature extraction and fusion: Early Fusion (EF) Structure, Siamese Structure, and Triplet Structure. The EF structure first fuses bi- temporal images through methods such as differencing or concatenation and then inputs them into a feature extraction network for feature analysis and change detection, e.g., Fully Convolutional Networks (FCN)\citep{ref27} and UNet\citep{ref28,ref29}. This structure is simple and easy to implement, but its major drawback lies in the pre-fusion of images before inputting into the network, disrupting the semantic information of different temporal phases and making it difficult to distinguish pseudo-changes caused by varying external factors. The Siamese structure employs two sub-networks to independently extract features from different temporal images and performs feature fusion and analysis during this process. The sub-networks are commonly composed of convolutional neural network (CNN)\citep{ref30,ref31,ref32} or vision transformer (ViT)\citep{ref33,ref34,ref35,ref36}. This structure greatly retains the geospatial features of the original images for difference analysis but may lack emphasis on change features. The Triplet structure\citep{ref37,ref38} enhances the identification of change areas by extracting features from three branches, analyzing both the geospatial features of the original images and their fused features. Such comprehensive analysis significantly improves the performance of change detection. However, current change detection methods still face challenges in accurately discerning change area boundaries, leading to a considerable amount of misdetections and omissions.

\par
\begin{figure}[!t]
	\centering
	\includegraphics[width=3.4in]{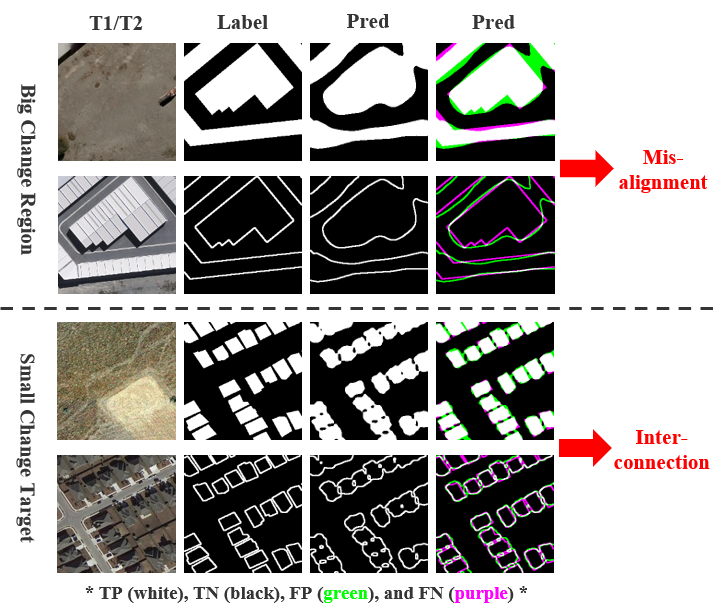}
	\caption{An illustration of the existing problems.}
	\label{fig1}
\end{figure}

 Although existing change detection methods have shown satisfactory performance, there is still much room for improvement. The accurate discrimination of edge details of change areas remains challenging. As shown in Fig. \ref{fig1}, discriminating boundaries of large change areas produces misalignment, and the boundaries of small change targets tend to become interconnected. Possible reasons for these challenges include insufficient focus on change features and a lack of constraints on the boundaries between change areas and the background. Additionally, during the commonly used pooling process in feature extraction, regional information may be smoothed or blurred, leading to distortion in data distribution and subsequently affecting boundary discrimination. Accurate discrimination of boundaries between change areas and the background is conducive to further improving the overall effectiveness of change detection tasks. Such improvement is essential for addressing more complex application scenarios.
\par To address the aforementioned issues, a novel change detection network, LRNet, based on a strategy of localization (change areas) followed by refinement (change edges), is proposed in this paper. LRNet consists of two stages: localization and refinement. In the localization stage, a three-branch encoder simultaneously extracts original image features and their differential features while interacting with each other to localize the position of each change area. To minimize information loss during feature extraction, the learnable optimal pooling (LOP) is proposed to replace the widely used max pooling. Additionally, LOP is trainable and participates in the overall optimization of the network. To effectively interact features from different branches and accurately localize change areas of various sizes, change alignment attention (C2A) and hierarchical change alignment module (HCA) are proposed. C2A is employed to interact change features from different branches, strengthening attention on change areas in a more reasonable way, weakening focus on the background, and generating the change alignment attention map (C2AM). The features with strengthened attention on change areas are then skip-connected to the corresponding level in the refinement stage to assist in the discrimination of change area edges. HCA hierarchically propagates C2AM in a densely connected manner to subsequent levels, aiding in the comprehensive discrimination of attention weights at different granularities. This ensures that attention maps generated for deep features are more representative and global, facilitating the accurate discrimination of change areas. In the refinement stage, the proposed Edge-Area Alignment module (E2A) corrects change areas and change edges of the localization results, providing a good start for subsequent edge refinement. Then, the decoder, combining with the C2A-enhanced differential features from the localization stage, refines change areas of different sizes and their edges. To address sample imbalance, this paper adopts a hybrid loss of Binary Cross-Entropy (BCE) and Intersection over Union (IOU) for supervised training optimization. Finally, we obtain the results with accurate discrimination of change areas and boundaries. The main contributions of this paper include the following points.

\begin{enumerate}[1)]
	
	\item{A novel change detection network, LRNet, is proposed based on a strategy of Localization-then-Refinement to address the issue of edge discrimination. The learnable optimal pooling (LOP) is proposed as a replacement for max pooling in feature extraction, aiming to reduce information loss and participate in network optimization.}
	
	\item{Change alignment attention (C2A) is proposed to interact features from different branches in a more reasonable way, enhancing attention on change areas. Additionally, hierarchical change alignment module (HCA) is proposed to hierarchically propagate the change alignment attention map (C2AM) in a densely connected manner to different levels, aiding in the accurate localization of change areas of various sizes.}
	
	\item{The Edge-Area Alignment module (E2A) is proposed to constrain change areas and change edges of the localization results. A hybrid loss of Binary Cross-Entropy (BCE) and Intersection over Union are adopted for supervised training to optimize the network.}
	
	\item{The proposed method achieves the best comprehensive evaluation metrics (F1 and OA, IOU) and the most accurate boundary discrimination results compared to 13 other state-of-the-art methods on two publicly available datasets.}
	
\end{enumerate}

\par The remainder of this paper is organized as follows. Section \ref{Related_Work} introduces the related work. Section \ref{Methodology} presents the proposed methodology in detail. Section \ref{Experiments} analyzes and discusses the experimental results. Section \ref{Conclusion} concludes this paper.

\section{Related work}
\label{Related_Work}

\subsection{DL-based change detection methods}
\label{DL_CD_2_1}

DL-based change detection methods can be roughly categorized into three types depending on feature extraction and fusion approaches: Early Fusion (EF), Siamese, and Triplet structures. The EF structure initially fuses bi-temporal images through operations like differencing or concatenation, and then inputs them into a feature extraction network for analysis and change detection. For instance, \citet{ref39} concatenated bi-temporal three-channel street-view images into a six-channel image, and then fed it into a fully convolutional network (FCN) for end-to-end inference. \citet{ref40} utilized an improved UNet++ for feature extraction and change detection on concatenated high-resolution satellite images. \citet{ref41} performed multivariate morphological reconstruction (MMR) on the differenced bi-temporal images before inputting them into the FCN-PP network for change feature extraction, aiming to suppress noise and pseudo-change areas introduced by direct differencing. \citet{ref42} employed FCN for feature extraction on concatenated very-high-resolution (VHR) RS images, improving change detection through a multi-task learning approach. The EF structure, fusing images before entering the network, affects the acquisition of deep semantic features from different temporal phases, making it challenging to discern pseudo-changes caused by varying external factors.

\par In response to the limitations of the EF structure, researchers have proposed the Siamese structure. The Siamese structure typically employs two identical sub-networks with shared weights to separately extract features from bi-temporal images, and perform feature fusion and analysis during this process. This structure significantly preserves object features of the original images for differential analysis. The sub-networks are commonly constructed using convolutional neural network (CNN) or visual transformer (ViT). \citet{ref43} first introduced the fully convolutional Siamese network for change detection, with FC-Siam-Conc and FC-Siam-Diff. Between the encoder and decoder, FC-Siam-Conc and FC-Siam-Diff employ concatenation and differencing, respectively, to fuse deep features extracted from bi-temporal images. \citet{ref44} utilized metric learning to discriminate changes based on the deep features of bi-temporal images. \citet{ref45} designed an image fusion network (IFN) based on the fully convolutional Siamese structure for feature extraction, incorporating a decoder with attention mechanisms for change detection. \citet{ref46} utilized a Siamese network to extract multi-scale feature groups from bi-temporal images, improving change detection accuracy by filtering out different pseudo-change information through geospatial position matching mechanism (PMM) and geospatial content reasoning mechanism (CRM). \citet{ref47} utilized a Siamese graph convolutional network (GCN) for feature extraction and employed a multi-scale fusion block for change analysis and detection. With the popularity of ViT, many researchers have introduced ViTs to extract global features and long-range contextual information from images. \citet{ref34} proposed a bi-temporal image transformer (BiT) to construct the temporal-spatial context. \citet{ref48} presented a visual change transformer (VcT) based on graph neural network (GNN) and ViT for change detection. \citet{ref49} captured local and global information of RS images using an adaptive ViT-based Siamese encoder and supervised model training and optimization through joint learning.

\par Recently, some scholars have proposed the Triplet structure by combining the characteristics and advantages of both EF and Siamese structures. The Triplet structure involves three branches that separately extract object features and their fused features from the original images for comprehensive analysis, strengthening the identification of changed areas and effectively improving change detection accuracy. \citet{ref38} employed a triplet feature extraction module to extract spatial-spectral features from multispectral images for detecting subtle changes. \citet{ref50} designed a three-stream network with a cross-stage feature fusion module for high-resolution RS image change detection. \citet{ref37} proposed a triple network incorporating joint multi-frequency difference feature enhancement (JM-DFE) module and full-scale Swin-Transformer for detecting changes of various scales. In this study, the proposed LRNet utilizes triplet encoders to simultaneously extract the original image features and their differential features, facilitating interactive localization of each changed area. To mitigate information loss during feature extraction, LOP is proposed as an alternative to the widely used max-pooling in CNNs.

\subsection{Attention mechanism}
\label{AM_2_2}

Attention is an indispensable cognitive ability in humans\citep{ref51}, enabling individuals to select useful information from a massive amount of accessed information, while ignoring other irrelevant or useless information. Inspired by the human attention mechanism, researchers have developed a series of attention mechanisms for neural networks through computer simulations, successfully applying them in natural language processing\citep{ref52,ref53,ref54}, computer vision\citep{ref55,ref56,ref57}, and other fields. The most common attention mechanisms in change detection tasks are channel attention and spatial attention.

\par The channel attention mechanism (CAM) is utilized to focus on the spectral channel features in images, with the importance of each channel encoded in the channel attention map, and its weights automatically calibrated during network training. Leveraging CAM, features representing channels relevant to change detection are enhanced, while channels irrelevant to the objective are suppressed. In this way, the network can learn to attend to channel information crucial for the target from multi-source heterogeneous data. Similarly, the spatial attention mechanism (SAM) is employed to focus on spatial pixels relevant to the goal. For instance, in change detection tasks, pixels in change areas are assigned higher weights in the attention map learned by SAM to enhance feature representation. Conversely, unchanged pixels are assigned lower weights to restrain their feature expression. Since CAM and SAM \citep{ref45} can enhance the expressive ability of a model from different perspectives, researchers combine them in serial or parallel ways to further improve the performance of the model. \citet{ref58} proposed an attention fusion module (AFM) based on CAM and SAM to adaptively capture change information in spatial and channel dimensions. \citet{ref59} introduced the co-attention module to handle building displacements in orthoimages and further explore their correlations. \citet{ref60} designed an adaptive attention fusion module to emphasize important change features and accelerate model convergence.

\par In addition to the application of CAM and SAM in change detection, some other specifically designed attention mechanisms have been proposed to address more complex problems. Self-attention is proposed to establish long-range dependencies between pixels, capturing global information to extract more comprehensive change features. \citet{ref61} designed a hierarchical self-attention augmented Laplacian pyramid expanding network for supervised change detection in high-resolution remote sensing images. \citet{ref62} proposed a temporal cross-attention module to model both temporal- and spatial-wise interactions, fully exploring complementary cues between bi-temporal features. In this paper, change alignment attention (C2A) is proposed to interact the change feature information from different branches, enhancing attention to change areas in a more reasonable manner. The hierarchical change alignment (HCA) module is proposed to jointly refine different-sized change areas with C2A, obtaining more accurate change detection results.

\section{Methodology}
\label{Methodology}

\subsection{Framework overview}

\begin{figure*}[!t]
	\centering
	\includegraphics[width=6in]{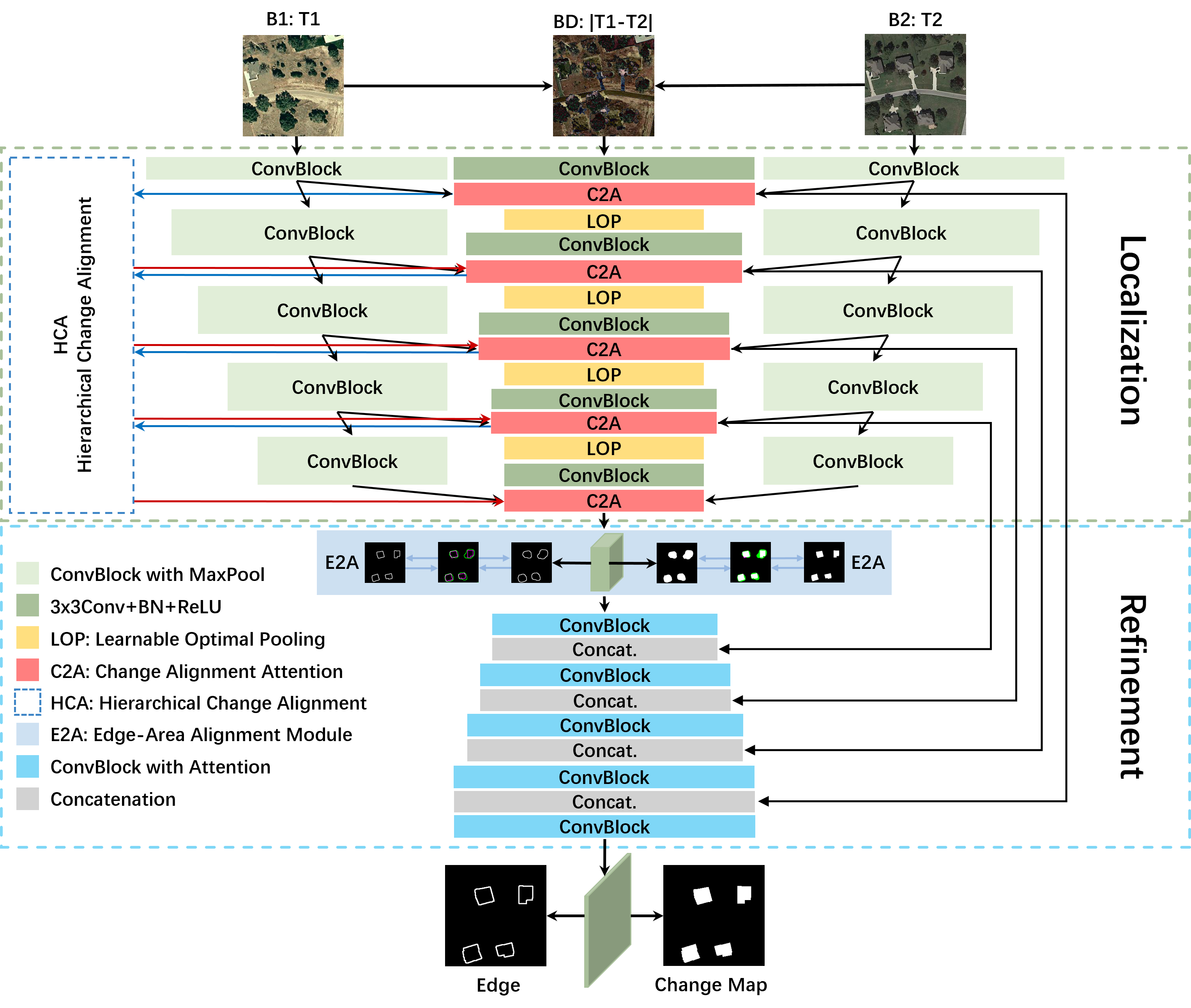}
	\caption{Flowchart of the LRNet network.}
	\label{fig2}
\end{figure*}

The proposed change detection network, LRNet, as illustrated in Fig. \ref{fig2}, consists of two stages: localization and refinement.

The localization stage involves a three-branch encoder, simultaneously extracting features from the original bi-temporal images and their differences, then interacting with them to pinpoint the location of each change area. The three branches are denoted as B1, B2, and BD, where B1 and B2 utilize pre-trained VGG16 from ImageNet for feature extraction from the original images. BD employs learnable optimal pooling (LOP) to replace the VGG16 from max-pooling as the backbone for difference feature extraction. LOP reduces information loss during feature extraction and participates in the optimization of the entire network, as detailed in Section \ref{LOP}. Let $I_{B1}$, $I_{B2}$, and $I_{BD}$ represent the inputs to branches B1, B2, and BD, respectively, where $I_{B1}$ and $I_{B2}$ are the original bi-temporal images. $I_{BD}$ is the difference image, and the relationship with $I_{B1}$ and $I_{B2}$ is described in Eq. \ref{eq1}.
\begin{equation}
	\label{eq1}
	I_{BD}= \lvert I_{B1} - I_{B2} \rvert
\end{equation}

Each branch consists of 5 convolutional blocks, each comprising several convolution layers, batch normalization layers, and ReLU layers. Branches B1 and B2 also include max-pooling layers. The output of the $y$-th ($y=1,2,3,4,5$) convolutional block in branch B$x (x=1,2,D)$ is denoted as $F_{Ly}^{Bx}$. After each convolutional block, we utilize the proposed change alignment attention (C2A) and hierarchical change alignment (HCA) modules to interact features from different branches, thereby accurately localizing change areas of different sizes. The feature transformation in C2A is described in Eq. \ref{eq2}. Details about C2A and HCA are provided in Sections \ref{C2A} and \ref{HCA}, respectively.
\begin{equation}
	\label{eq2}
	F_{Ly}^{C2A},C2AM_{Final}^{Ly}={C2A}_{Ly}\left(F_{Ly}^{B1},F_{Ly}^{B2},F_{Ly}^{BD},C2AM_{Pre}^{Ly}\right)
\end{equation}
Where ${C2A}_{Ly}$ represents the $y$-th ($y=1,2,3,4,5$) C2A module, taking inputs from $F_{Ly}^{B1}$,$F_{Ly}^{B2}$,$F_{Ly}^{BD}$,$C2AM_{Pre}^{Ly}$, and producing outputs $F_{Ly}^{C2A}$ and $C2AM_{Final}^{Ly}$. $C2AM_{Pre}^{Ly}$ is the change-aligned attention map output by the corresponding level HCA, while $C2AM_{Final}^{Ly}$ is the change-aligned attention map input to the corresponding level HCA. $F_{Ly}^{C2A}$ represents the enhanced feature aligned with change areas, which is further exploited for precise localization in subsequent layers and is connected to the refinement stage via skip connections to assist in refining change targets of different sizes. In particular, since $C2AM_{Pre}^{Ly}$ serves as the fused attention of previous stages as well as the first layer C2A and HCA do not have a previous stage, there does not exist $C2AM_{Pre}^{L1}$. So the feature transformation of the first layer C2A is shown in Eq. \ref{eq2_1}. Finally, the feature output from the localization stage is denoted as $F_{L5}^{C2A}$, serving as the input to the refinement stage.
\begin{equation}
	\label{eq2_1}
	F_{L1}^{C2A},C2AM_{Final}^{L1}={C2A}_{L1}\left(F_{L1}^{B1},F_{L1}^{B2},F_{L1}^{BD}\right)
\end{equation}

\par The refinement stage starts with constraining and refining the change areas and edges obtained from the localization stage using the proposed edge-area alignment (E2A) module, providing a solid starting point for subsequent edge refinement. Details of E2A are discussed in Section \ref{e2a}. Corresponding to the localization stage, the refinement stage comprises five convolutional blocks, each consisting of several convolution layers, transpose convolution layers, batch normalization layers, and ReLU layers. Each block refines different-sized change areas and their edges from deep to shallow in combination with difference features enhanced by C2A from the localization stage. Let $F_{Rz}$ represent the output of the $z$-th ($z=5,4,3,2,1$) convolutional block in the refinement stage, ultimately resulting in the change intensity map $F_{R1}$. Threshold segmentation using the sigmoid function yields accurately discriminated change areas and edges. The structure and role of each module are described in detail below.

\subsection{Learnable optimal pooling}
\label{LOP}

To address information loss during feature extraction, the discussion is primarily about the pooling layer in CNN. The pooling idea originates from the visual mechanism, which is the process of abstracting information. It can enlarge receptive fields while reducing feature dimensions, decreasing network parameters with lower optimization difficulty. Max pooling, a commonly used technique, highlights texture features but smooths out non-maximal features within the kernel region. Additionally, some scholars have employed dilated convolutions to emulate pooling layers; however, this approach often results in information loss as the internal details of dilated pixels within convolution kernel are disregarded.

\begin{figure}[!t]
	\centering
	\includegraphics[width=3.2in]{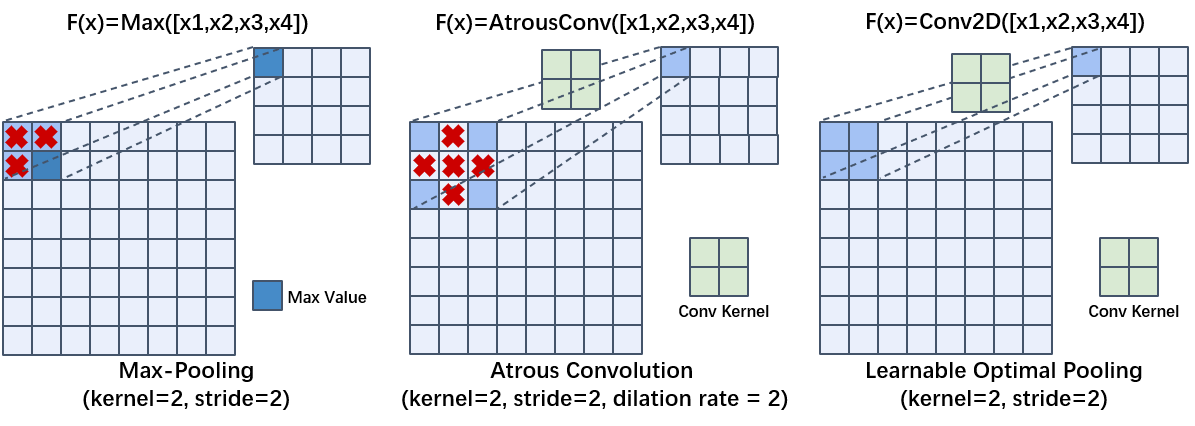}
	\caption{Schematic diagram of learnable optimal pooling.}
	\label{fig3}
\end{figure}

As shown in Fig. \ref{fig3}, to address aforementioned issues, we propose learnable optimal pooling (LOP). Unlike dilated convolutions, LOP adopts a dense convolution kernel strategy, utilizing convolutional learning to compute aggregated information within the kernel region rather than relying on simple statistical values like the maximum. This pooling strategy ensures that the kernels cover all feature regions, reducing information loss. Meanwhile, the convolutional parameters are used to learn the regional aggregation information, avoiding the information from being smoothed excessively and ensuring that the aggregated features remain representative. LOP can reduce information loss during feature extraction and ensure comprehensive key change features for assisted localization and refinement, which in turn improves the detection performance of the network. The BD branch depicted in Fig. \ref{fig2} contains four LOP modules, each LOP performs feature aggregation on the output of C2A, enlarging the receptive field while reducing feature scale. The transformation process of LOP is illustrated in Eq. \ref{eq3}.
\begin{equation}
	\label{eq3}
	F_{Ly}^{LOP}=LOP_{Ly}(F_{Ly}^{C2A})
\end{equation}
Here, $LOP_{Ly}$ represents the $y$-th ($y=1,2,3,4$) LOP module, taking the output $F_{Ly}^{C2A}$ of the $y$-th C2A as input and producing $F_{Ly}^{C2A}$ as the input for the next convolutional block.

\subsection{Change alignment attention}
\label{C2A}

\begin{figure*}[!t]
	\centering
	\includegraphics[width=5.2in]{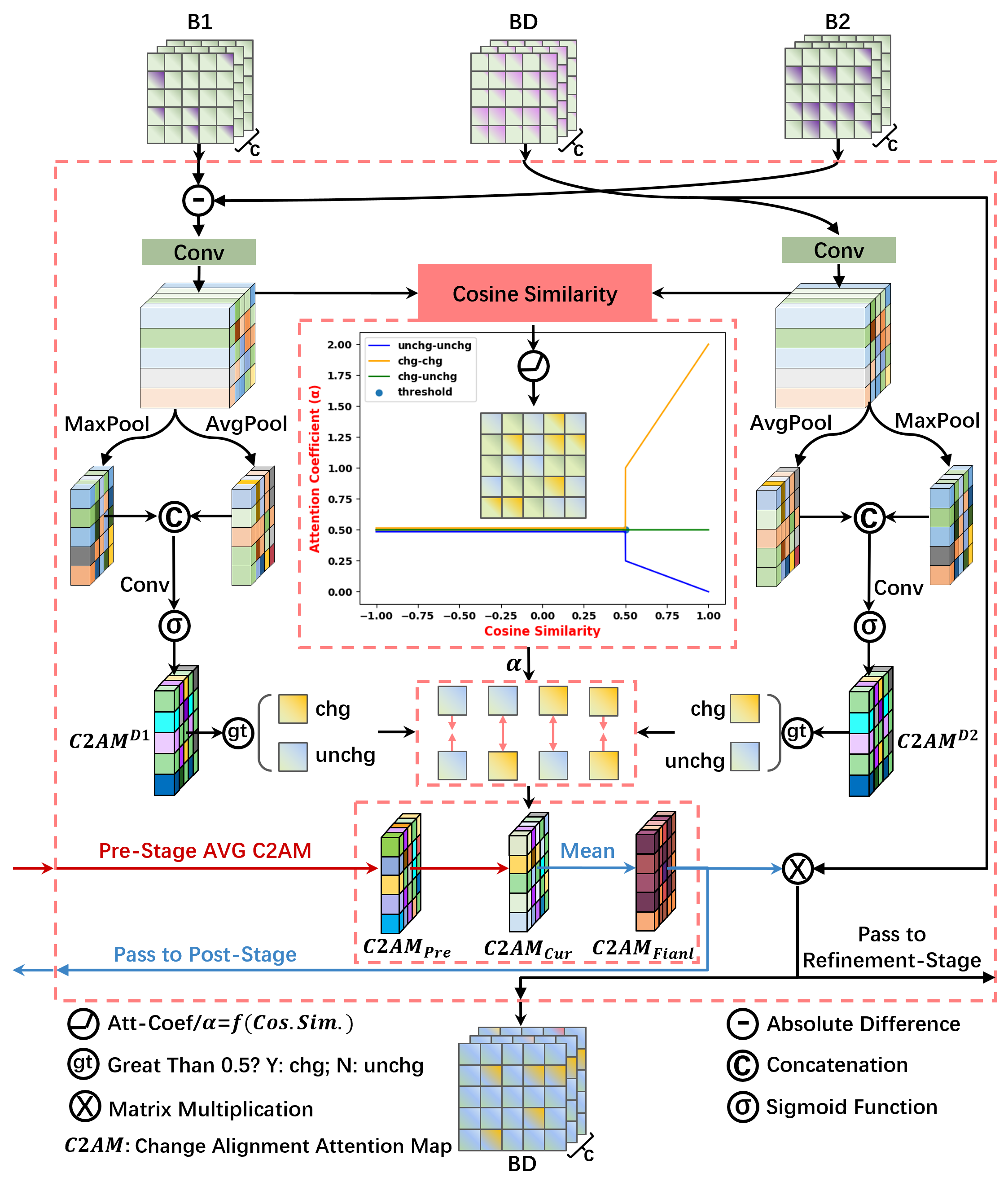}
	\caption{Structure of the change alignment attention module.}
	\label{fig4}
\end{figure*}

To facilitate effective interaction among features from different branches and accurately locate change areas of various sizes, the change alignment attention (C2A) and hierarchical change alignment (HCA) modules are designed. The processing of C2A is illustrated in Fig. \ref{fig4}. The inputs consist of features from the three branches, denoted as $F_{Ly}^{B1}$,$F_{Ly}^{B2}$,$F_{Ly}^{BD}$, and the change alignment attention map from the corresponding HCA module, denoted as $C2AM_{Pre-y}$. Initially, features from branches B1 and B2 undergo differencing to highlight changes, followed by convolution with the features from branch BD to obtain the original change features, denoted as $F_{Ly}^{D1}$ and $F_{Ly}^{D2}$, as shown in Eq. \ref{eq4}.
\begin{equation}
	\label{eq4}
	\begin{cases} 
		F_{Ly}^{D1}=&Conv(\lvert F_{Ly}^{B1}-F_{Ly}^{B2} \rvert) \\
		F_{Ly}^{D2}=&Conv(F_{Ly}^{BD})
	\end{cases}
\end{equation}

The original change features, $F_{Ly}^{D1}$ and $F_{Ly}^{D2}$, undergo information aggregation through max-pooling and average-pooling. Subsequently, the pooled information is concatenated and processed through convolutional learning to obtain the preliminary attention weights, denoted as $C2AM_{Ly}^{D1}$ and $C2AM_{Ly}^{D2}$, as shown in Eq. \ref{eq5}.
\begin{equation}
	\label{eq5}
	C2AM_{Ly}^{Di}=\sigma(Conv([MaxPool(F_{Ly}^{Di});AvgPool(F_{Ly}^{Di})]))
\end{equation}
Where, $i=1,2$ , refers to the corresponding branch, $[;]$ denotes the concatenation operation, and $\sigma(\cdot)$ represents the sigmoid function.

Due to their different sources, $F_{Ly}^{D1}$ and $F_{Ly}^{D2}$ contain different and complementary information. The former emphasizes more on change features highlighted by direct differencing, while the latter contains more differential information from deep features of the bi-temporal images. By analyzing the similarity of the original change features, weights are assigned to the attention to enhance consistent information. Areas consistent in both are assigned higher coefficients for change category and lower coefficients for unchanged category. First, we threshold the attention weight map to obtain a flag indicating whether each pixel represents change, using a common threshold of 0.5, as shown in Eq. \ref{eq6}.
\begin{equation}
	\label{eq6}
	{flag_p}_{Ly}^{Di}=
	\begin{cases} 
		chg, &{C2AM_p}_{Ly}^{Di}\geq0.5 \\
		unchg, &{C2AM_p}_{Ly}^{Di}\textless0.5
	\end{cases}
	,p\in[1,h\times w]
\end{equation}
Where, $h$ and $w$ represent the height and width of $C2AM_{Ly}^{Di}$, respectively, ${flag_p}_{Ly}^{Di}$ denotes the flag corresponding to each pixel in ${C2AM_p}_{Ly}^{Di}$. The combination of ${flag_p}_{Ly}^{D1}$ and ${flag_p}_{Ly}^{D2}$ yields four possible cases: $c1.$ change vs. change, $c2.$ unchanged vs. unchanged, $c3.$ change vs. unchanged, and $c4.$ unchanged vs. change. The last two cases can be regarded as the same.

The cosine similarity ($sim$) of the original change features is calculated, and the attention weight coefficient $\alpha$ is obtained in accordance with the similarity threshold $T$ as well as the weighting rule. The corresponding formulas are shown in Equations \ref{eq7} and \ref{eq8}.
\begin{equation}
	\label{eq7}
	sim_{p}^{Ly}=\frac{{F_p}_{Ly}^{D1}\cdot{F_p}_{Ly}^{D2}}{\left\|{F_p}_{Ly}^{D1}\right\|\left\|{F_p}_{Ly}^{D2}\right\|}
\end{equation}
\begin{equation}
	\label{eq8}
	\alpha_{p}^{Ly}=
	\begin{cases} 
		T, &sim_{p}^{Ly}\leq T \\
		2\times sim_{p}^{Ly}, &sim_{p}^{Ly}>T and ({flag_p}_{Ly}^{D1},{flag_p}_{Ly}^{D2}) \in \{c1\} \\
		1-sim_{p}^{Ly}, &sim_{p}^{Ly}>T and ({flag_p}_{Ly}^{D1},{flag_p}_{Ly}^{D2}) \in \{c2\} \\
		T, &sim_{p}^{Ly}>T and ({flag_p}_{Ly}^{D1},{flag_p}_{Ly}^{D2}) \in \{c3,c4\}
	\end{cases}
\end{equation}
Here, $p\in[1,h\times w]$, $h$ and $w$ represent the height and width of the corresponding feature map, respectively. In the experiments, $T$ is set to 0.5.

According to $\alpha$, the weighted fusion of $C2AM_{Ly}^{D1}$ and $C2AM_{Ly}^{D2}$ yields an attention weight map $C2AM_{Cur}^{Ly}$ that enhances the representation of change category and suppresses the representation of unchanged category. Considering the different levels of attention information, the final change-aligned attention weight map $C2AM_{Final}^{Ly}$ for current stage is obtained by averaging with the previously fused attention $C2AM_{Pre}^{Ly}$, as shown in Eq. \ref{eq9}.
\begin{equation}
	\label{eq9}
	\begin{cases} 
		C2AM_{Cur}^{Ly}=\alpha\times (C2AM_{Ly}^{D1}+C2AM_{Ly}^{D2})/2 \\
		C2AM_{Final}^{Ly}=(C2AM_{Cur}^{Ly}+C2AM_{Pre}^{Ly})/2
	\end{cases}
\end{equation}
Where $C2AM_{Pre}^{Ly}$ represents the output of the corresponding level HCA module, and the obtained weight map $C2AM_{Final}^{Ly}$, serving as the input for the corresponding level HCA module, are used for the fusion of different level attentions in the following stage. The product of $C2AM_{Final}^{Ly}$ and the input feature $F_{Ly}^{BD}$ yields the new feature $F_{Ly}^{C2A}$, enhancing the alignment of change areas. $F_{Ly}^{C2A}$ is utilized both for further exploration of change features to aid in localization in the BD branch and for skip connection to the refinement stage to assist in the refinement of change targets of different sizes. Overall, C2A facilitates the interaction of change feature information from different branches, aligning and enhancing change areas in a more rational way, and reducing the focus on the background.

\subsection{Hierarchical change alignment}
\label{HCA}

\begin{figure}[!t]
	\centering
	\includegraphics[width=3.2in]{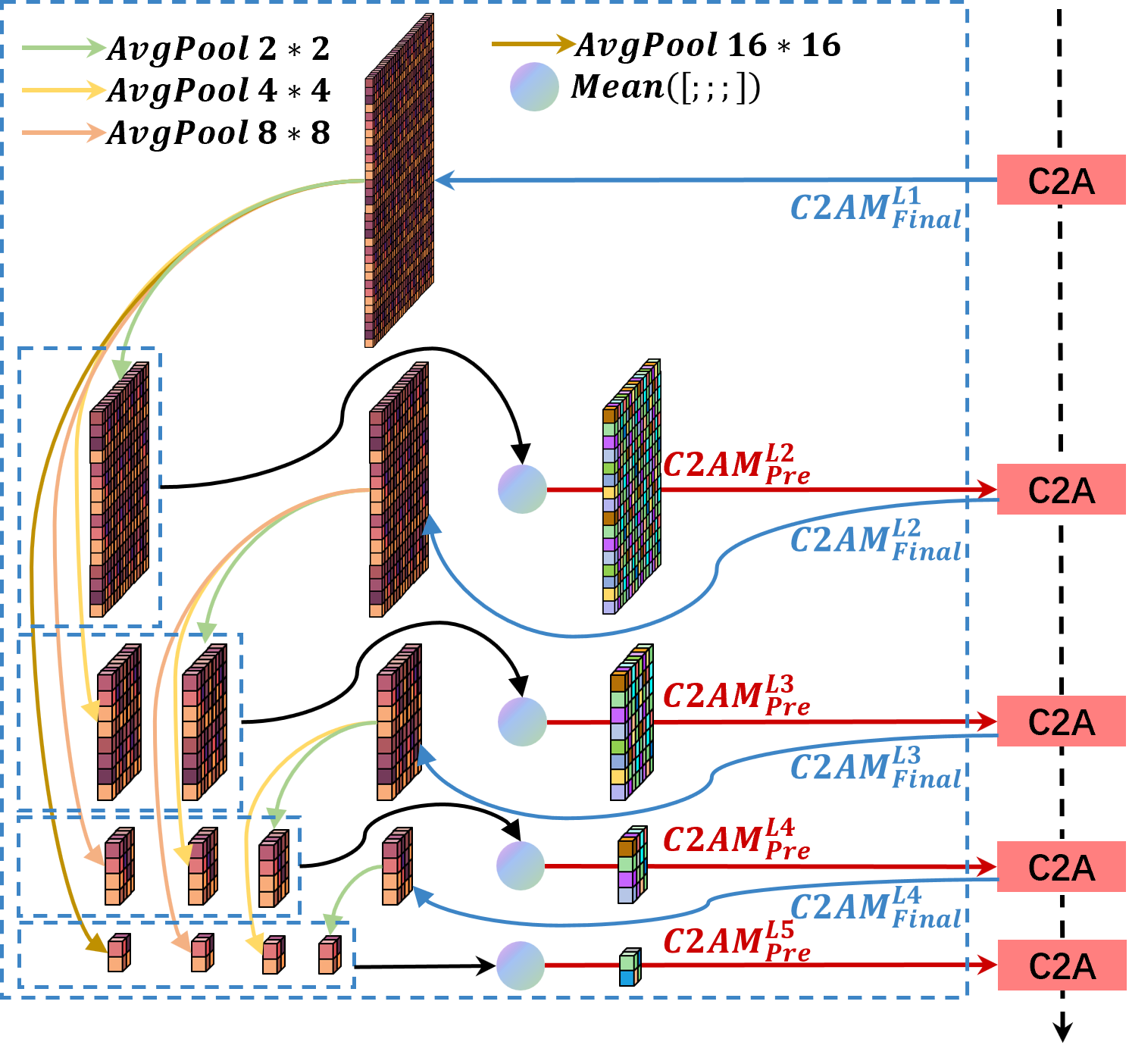}
	\caption{Structure of the hierarchical change alignment module.}
	\label{fig5}
\end{figure}

In the network architecture, features at different levels exhibit varying granularity and receptive fields, which can be utilized for identifying targets of different sizes. To effectively synthesize change-aligned attention from different granularities, the hierarchical change alignment module (HCA) is proposed. The structure of HCA is depicted in Fig. \ref{fig5}. HCA propagates the change alignment attention maps (C2AM) outputted from different level C2A to subsequent levels hierarchically in a densely connected manner, facilitating comprehensive discrimination of attention weights at different granularities. It ensures that attention weight generation for deep features is more representative, which is conducive to the identification of change targets of various sizes.

HCA consists of 5 layers corresponding to C2A. Specifically, the first layer of HCA has only input, represented by $C2AM_{Final}^{L1}$; the last layer of HCA has only output, represented by $C2AM_{Pre}^{L5}$. Other layers have both input and output, where the input is the change-aligned attention map from the corresponding layer of C2A, denoted as $C2AM_{Final}^{Ly}$, and the output is the fused attention from previous layers, denoted as $C2AM_{Pre}^{Ly}$. Taking $C2AM_{Pre}^{L3}$ as an example, the processing of HCA is demonstrated in Eq. \ref{eq10}.
\begin{equation}
	\label{eq10}
	\begin{split}
		C2AM_{Pre}^{L3}=Mean[AvgPool4(C2AM_{Final}^{L1});\\
		AvgPool2(C2AM_{Final}^{L2})]
	\end{split}
\end{equation}

In the equation, $[;]$ denotes concatenation operation, $AvgPool4(\cdot)$ and $AvgPool2(\cdot)$ represent average pooling with kernel size of 4*4 and 2*2, respectively. Following the same procedure, we obtain $C2AM_{Final}^{L2}$, $C2AM_{Final}^{L4}$, and $C2AM_{Final}^{L5}$, which all contain fused information of change-aligned attention from different levels in the preceding layer, facilitating the localization and refinement of change targets of various sizes. The output of the last layer HCA, $C2AM_{Final}^{L5}$, serves as input to the final layer of C2A, combined with change features from the three branches to obtain accurately localized change areas in $F_{L5}^{C2A}$. $F_{L5}^{C2A}$ serves as input to the refinement stage, proceeding from deep to shallow to refine change areas of different sizes and their edges. The structure and modules of the refinement stage are discussed next.

\subsection{Edge-area alignment module}
\label{e2a}
\begin{figure}[!t]
	\centering
	\includegraphics[width=3.2in]{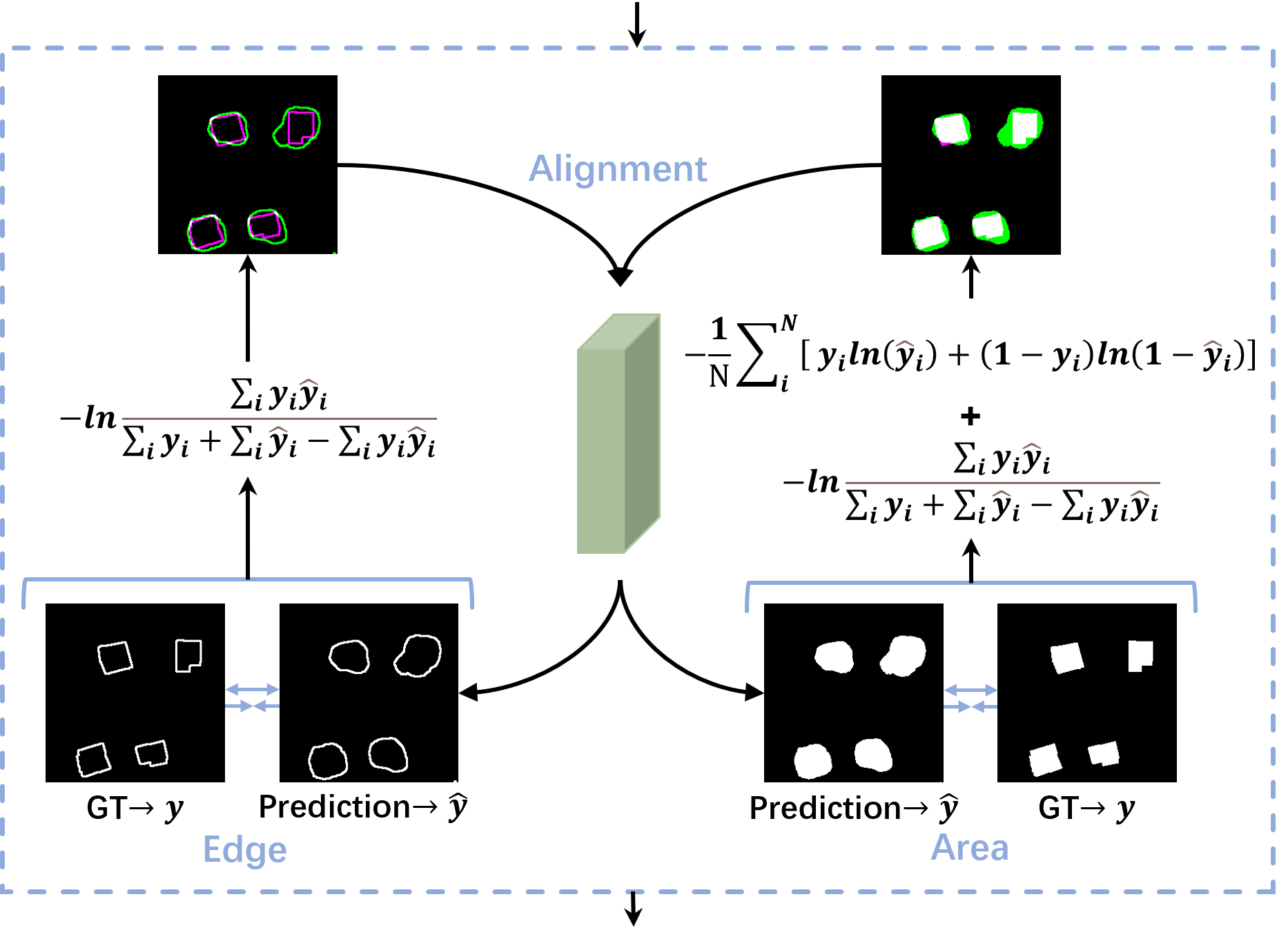}
	\caption{Schematic diagram of the edge-area alignment module.}
	\label{fig6}
\end{figure}

To provide a solid foundation for the refinement stage, the edge-area alignment module (E2A) is designed to constrain and rectify the change areas and edges identified in the localization stage. The procedure of E2A is illustrated in Fig. \ref{fig6}. Firstly, the output from the localization stage is processed using convolution and sigmoid function to get the inference result of change areas. Subsequently, the Canny algorithm\citep{ref63} is applied on the change areas to get the edge detection results. Comparing the inference result with the ground truth, different loss functions are employed for supervision. Inspired by lane detection in autonomous driving\citep{ref64}, it is more stable to use intersection over union (IOU) for evaluation and constraints when the target proportion is significantly smaller than the background, which allows us to focus primarily on the less-targeted category. Specifically, we employ IOU loss function to constrain the detection results of change edges, while for change areas, we combine binary cross-entropy loss function with IOU loss function for joint constraint. Equations \ref{eq11} and \ref{eq12} are the formulas for binary cross-entropy loss function and IOU loss function, respectively.
\begin{equation}
	\label{eq11}
	L_{BCE}=-\frac{1}{N}\sum\nolimits_i^N\left[y_iln\left(\hat{y_i}\right)+\left(1-y_i\right)ln\left(1-\hat{y_i}\right)\right]
\end{equation}
\begin{equation}
	\label{eq12}
	L_{IOU}=-ln\frac{\sum_i{y_i\hat{y_i}}}{\sum_i{y_i}+\sum_i{\hat{y_i}}-\sum_i{y_i\hat{y_i}}}
\end{equation}
Where $y_i$ represents the ground truth, $\hat{y_i}$ denotes the inference result, and $N$ refers to the total number of pixels.

The constraints for edge, area, and overall are shown in Eq. \ref{eq13}.
\begin{equation}
	\label{eq13}
	\begin{cases} 
		L_{area}=L_{BCE}+L_{IOU}\\
		L_{edge}=L_{IOU}\\
		L_{total}=L_{area}+L_{edge}
	\end{cases}
\end{equation}

By imposing constraints on the changed edges and areas between the localization and refinement stage, the inference values are brought closer to the ground truth, laying a solid foundation for further refinement of the change details. Additionally, at the end of the refinement stage, we supervise the training and optimization of the model by constraining the changed edges and areas in the same manner.

\subsection{Refinement stage}
\label{section_3_6}

As illustrated in Fig. \ref{fig2}, the refinement stage utilizes the E2A module to process the localization results, yielding deep features corrected for changed areas and edges. Considering that deep features contain more abstract semantic information while shallow features contain more detailed change information, the refinement stage integrates different levels of features passed from C2A, from deep to shallow and from center to edge, to refine the delineation of changed areas and edges.

Corresponding to the localization stage, the refinement stage comprises five convolutional blocks, each consisting of several convolution layers, transpose convolution layers, batch normalization layers, and ReLU layers. The features from the first four convolutional blocks are fused with the change-enhanced features by the corresponding C2A module, refining the edges of different-sized changed areas. CAM is embedded in the concatenation operation to assist in feature fusion and filtering of redundant information, and SAM is embedded to enhance the change features before enlarging the resolution. Progressively refining the changed features from deep to shallow, the final convolutional block generates the change intensity map $F_{R1}$. Utilizing a sigmoid function on $F_{R1}$ yields the change detection results with accurate area and edge discrimination.

\section{Experiments}
\label{Experiments}
In this section, we first introduce the datasets, experimental setting, and evaluation metrics in turn. And then we provide a detailed analysis of the comparison experiments. Finally, we discuss the effectiveness of the proposed modules and loss functions, as well as the complexity of the model.

\subsection{Data description}
The experiments are conducted on two publicly available change detection datasets, namely, the LEVIR-CD and WHU-CD datasets.

1) LEVIR-CD dataset\citep{ref65}: This dataset consists of 637 pairs of high-resolution remote sensing images, each with a size of 1024*1024 and a spatial resolution of 0.5 m/pixel. To facilitate training and reduce memory burden, the original images are cropped into new pairs of size 256*256 without overlapping, and finally 10,192 pairs of samples are obtained. In the experiments, the dataset is randomly divided according to the default official split ratio of 7:1:2. The training, validation, and test sets contain 7120, 1024, and 2048 pairs of samples, respectively

2) WHU-CD dataset\citep{ref66}: This dataset comprises one pair of high-resolution remote sensing images, each with a size of 32,507*15,354 and a spatial resolution of 0.3 m/pixel. Similarly, the original images are cropped into blocks of size 256*256, resulting in 7620 pairs of samples. The dataset is randomly divided into training, validation, and test sets with a ratio of 8:1:1, containing 6096, 762, and 762 pairs of samples, respectively.

\begin{figure*}[!t]
	\centering
	\includegraphics[width=6.8in]{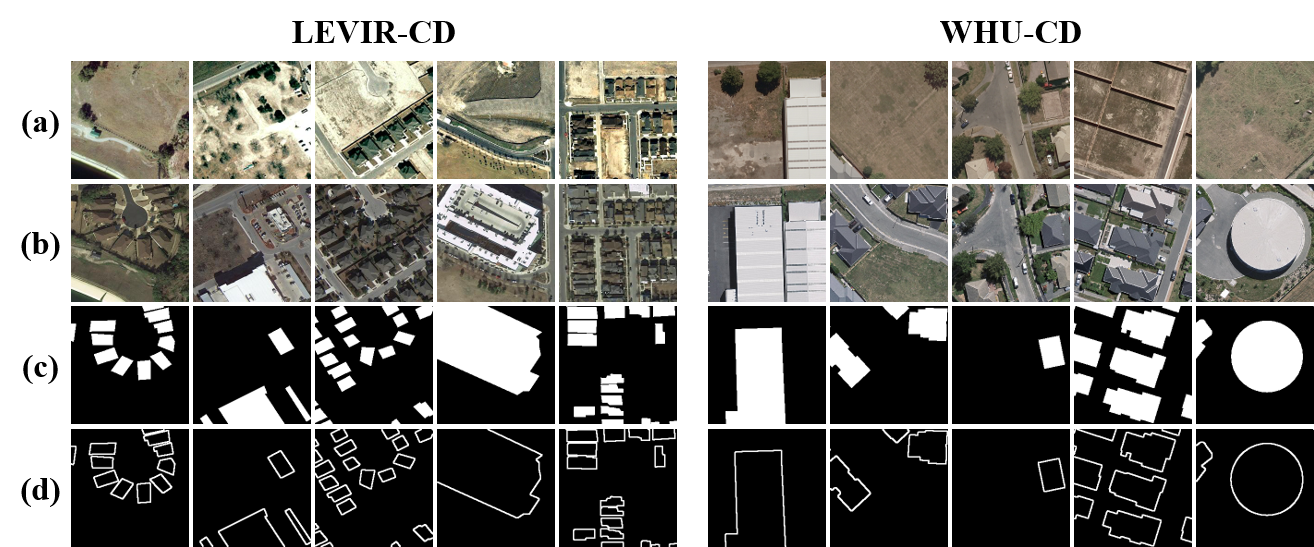}
	\caption{Some samples from the LEVIR-CD and WHU-CD datasets. (a) T1 image (b) T2 image (c) Label area (d) Label edge.}
	\label{fig7}
\end{figure*}

Some samples from the LEVIR-CD and WHU-CD datasets are shown in Fig. \ref{fig7}.

\subsection{Experimental setting}
In the experiments, the proposed LRNet is compared with 13 other state-of-the-art change detection methods. As shown in Table \ref{tab1}, CNN-based methods include FC-EF, FC-Siam-Conc, FC-Siam-Diff, DASNet, SNUNet, DSIFN, DESSN, DMINet, GeSANet, and TCD-Net; ViT-based methods include BIT, VcT, and EATDer. FC-EF and TCD-Net are constructed based on the EF structure and the triplet structure, respectively, while the other compared methods are based on the Siamese structure.

\begin{table*}[!t]
	\centering
	\caption{Description of compared methods.}
	\label{tab1}
	\begin{tabular}{cccc}
		\toprule
		Model & Publication Year & Based & Encoder Type \\
		\midrule
		FC-EF\citep{ref43} & 2018 & CNN & Early Fusion \\ 
		FC-Siam-Conc\citep{ref43} & 2018 & CNN & Siamese \\ 
		FC-Siam-Diff\citep{ref43} & 2018 & CNN & Siamese \\ 
		DSIFN\citep{ref45} & 2020 & CNN & Siamese \\ 
		DASNet\citep{ref44} & 2021 & CNN & Siamese \\ 
		SNUNet\citep{ref67} & 2022 & CNN & Siamese \\ 
		DESSN\citep{ref68} & 2022 & CNN & Siamese \\ 
		BIT\citep{ref34} & 2022 & ViT & Siamese \\ 
		VcT\citep{ref48} & 2023 & ViT & Siamese \\ 
		DMINet\citep{ref30} & 2023 & CNN & Siamese \\ 
		GeSANet\citep{ref46} & 2023 & CNN & Siamese \\ 
		TCD-Net\citep{ref37} & 2023 & CNN & Triplet \\ 
		EATDer\citep{ref49} & 2024 & ViT & Siamese \\ 
		\bottomrule
	\end{tabular}
\end{table*}

All models are trained and tested on an NVIDIA GeForce RTX 3090 GPU with 24 GB of memory and are implemented using Pytorch. In the experiments, the Adam optimization algorithm\citep{ref69} with an initial learning rate of 1e-4 is used to update the weights of the model. LRNet is trained for 200 epochs on each dataset with a batch size of 16. The implementation details of the compared methods are consistent with their respective papers.

\subsection{Evaluation metrics}

In the experiments, five evaluation metrics are adopted to quantitatively evaluate the change detection performance of different methods: Precision (Pre), Recall (Rec), F1-score (F1), Intersection over Union (IOU), and Overall Accuracy (OA). Pre and Rec can reflect the classification accuracy of the model from different perspectives. Changed pixels are defined as positive samples, while unchanged pixels are negative samples. Pre refers to the proportion of correctly predicted positive samples among all samples predicted as positive. Rec refers to the proportion of correctly predicted positive samples among all positive samples in the dataset. For the change detection task, a higher Pre value indicates fewer false detections, while a higher Rec value indicates fewer missed detections.

F1, IOU, and OA can reflect the overall performance of the model. F1 is the harmonic average of Pre and Rec. IOU measures the overlap between the predicted values and ground truth for positive samples, where a larger value indicates a closer match to the ground truth. OA refers to the proportion of correctly predicted pixels among all pixels. The definitions of these metrics are shown in Eq. \ref{eq14}.
\begin{equation}
	\label{eq14}
	\begin{cases}
		Pre={TP}/(TP+FP)\\
		Rec={TP}/(TP+FN)\\
		F1=(2*Pre*Rec)/(Pre+Rec)\\
		IOU={TP}/(TP+FP+FN)\\
		OA=(TP+TN)/(TP+TN+FP+FN)
	\end{cases}
\end{equation}
Where $TP$, $TN$, $FP$, and $FN$ represent the numbers of true positives, true negatives, false positives, and false negatives, respectively.

To evaluate the detection performance of the model on change edges, we not only compare the detection metrics of change areas but also further measure the metrics of change edges: $Pre_{Edge}$, $Rec_{Edge}$, $F1_{Edge}$, $IOU_{Edge}$. These metrics are computed in the same way as the change area metrics but focus on the change edges instead.

\subsection{Comparison with SOTA methods}
\label{sec_4_4}

1) Comparison experiments on the LEVIR-CD dataset: Table \ref{tab2} presents the quantitative results of each model on the LEVIR-CD dataset, where the best results are highlighted in bold. The proposed LRNet achieves optimal results in the overall evaluation metrics of change areas: OA (99.10\%), F1 (91.08\%), and IOU (83.63\%). It is slightly lower than the other compared methods in Pre (92.19\%) and Rec (90.00\%). Among them, FC-Siam-Conc exhibits the highest Rec (96.44\%), but its Pre (48.56\%) is significantly lower than other SOTA methods and LRNet. The oversimplified structure of FC-Siam-Conc struggles to differentiate between real and pseudo-changes, resulting in a high Rec but low Pre scenario, where many pseudo-changes or unchanged objects are misclassified as changes. TCD-Net achieves the best Pre (92.70\%) by effectively utilizing three branches to extract features from original bi-temporal images and their difference features for change detection. However, it tends to miss some ambiguous changed objects, reflected in its lower Rec (88.16\%). Considering both Pre and Rec, LRNet achieves the best F1, with improvements ranging from 0.71\% to 39.8\% compared to other models. For the discrimination of change edges, LRNet achieves the best values in all four evaluation metrics: $Pre_{Edge}$ (72.05\%), $Rec_{Edge}$ (67.96\%), $F1_{Edge}$ (70.95\%), and $IOU_{Edge}$ (54.78\%), with improvements of 1.27\%, 0.67\%, 2.45\%, and 2.69\%, respectively, compared to the suboptimal values. These quantitative results demonstrate the outstanding change detection performance of LRNet, especially improving the discrimination of change edges.

\begin{table*}[!t]
	\centering
	\caption{Evaluation metrics of each change detection model on the LEVIR-CD dataset.}
	\label{tab2}
	
	\begin{tabular}{cccccccccc}
		\toprule
		Model & OA(\%) & Pre(\%) & Rec(\%) & F1(\%) & IOU(\%) & $Pre_{Edge}$(\%) & $Rec_{Edge}$(\%) & $F1_{Edge}$(\%) & $IOU_{Edge}$(\%) \\
		\midrule
		FC-EF & 90.72 & 35.01 & 95.79 & 51.28 & 34.49 & 5.11 & 9.42 & 6.63 & 3.43 \\ 
		FC-Siam-Conc & 94.61 & 48.56 & \textbf{96.44} & 64.59 & 47.71 & 9.01 & 11.59 & 10.14 & 5.34 \\ 
		FC-Siam-Diff & 96.06 & 57.09 & 91.83 & 70.41 & 54.34 & 13.23 & 11.86 & 12.51 & 6.67 \\ 
		DASNet & 98.43 & 82.81 & 87.49 & 85.09 & 74.05 & 14.83 & 16.19 & 15.48 & 8.39 \\ 
		SNUNet & 98.92 & 91.23 & 87.29 & 89.22 & 80.54 & 67.33 & 65.61 & 66.46 & 49.77 \\ 
		DSIFN & 98.95 & 89.43 & 89.52 & 89.48 & 80.97 & 68.36 & 61.87 & 64.95 & 48.10 \\ 
		DESSN & 99.01 & 92.59 & 87.66 & 90.06 & 81.92 & 68.92 & 67.29 & 68.10 & 51.63 \\ 
		BIT & 98.84 & 91.18 & 85.59 & 88.30 & 79.04 & 63.20 & 63.45 & 63.32 & 46.33 \\ 
		VcT & 98.90 & 89.39 & 88.96 & 89.18 & 80.47 & 65.12 & 62.70 & 63.89 & 46.94 \\ 
		DMINet & 98.94 & 91.05 & 87.84 & 89.42 & 80.86 & 66.91 & 66.20 & 66.55 & 49.87 \\ 
		GeSANet & 99.03 & 91.64 & 89.04 & 90.32 & 82.35 & 69.95 & 65.90 & 67.87 & 51.36 \\ 
		EATDer & 98.94 & 89.83 & 89.36 & 89.59 & 81.15 & 66.08 & 66.26 & 66.17 & 49.44 \\ 
		TCD-Net & 99.04 & \textbf{92.70} & 88.16 & 90.37 & 82.43 & 70.78 & 66.36 & 68.50 & 52.09 \\ 
		LRNet & \textbf{99.10} & 92.19 & 90.00 & \textbf{91.08} & \textbf{83.63} & \textbf{72.05} & \textbf{67.96} & \textbf{70.95} & \textbf{54.78} \\ 
		\bottomrule
	\end{tabular}
	
\end{table*}

\begin{figure*}[!t]
	\centering
	\includegraphics[width=6.8in]{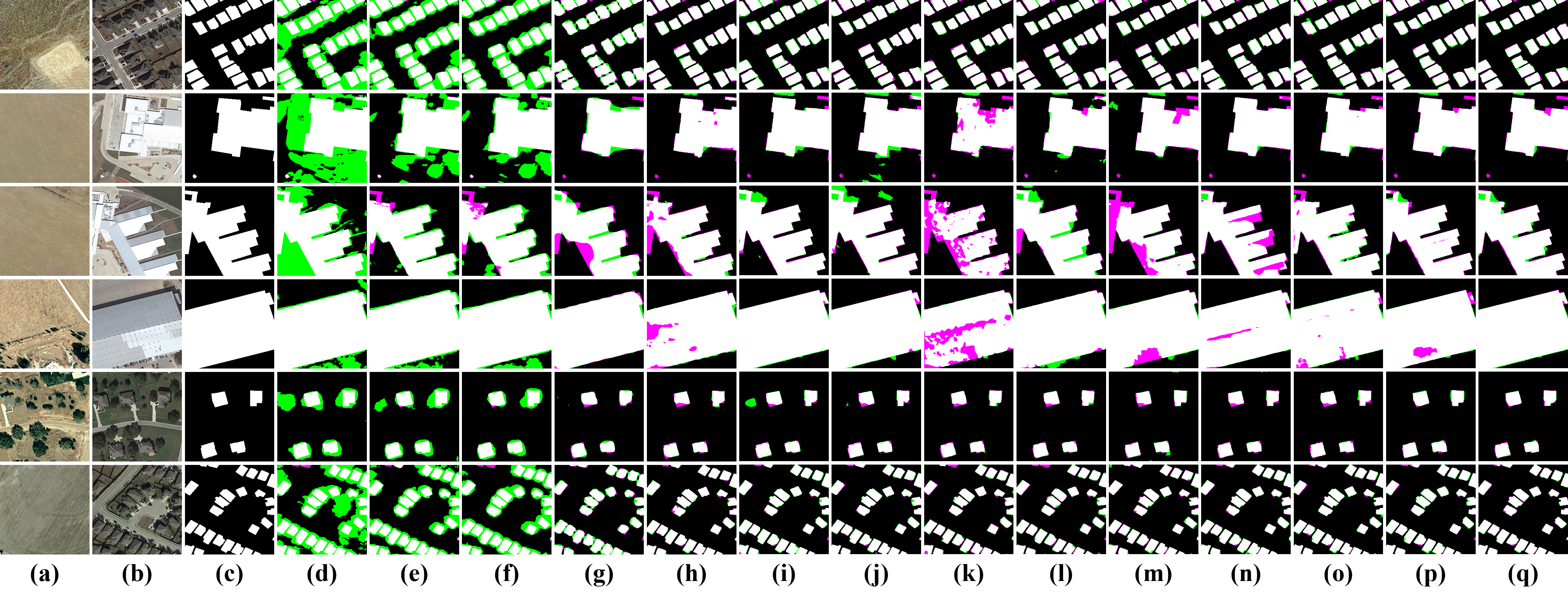}
	\caption{Visualization of change detection results on the LEVIR-CD dataset (a) T1 image; (b) T2 image; (c) Label; (d) FC-EF; (e) FC-Siam-Conc; (f) FC-Siam-Diff; (g) DASNet; (h) SNUNet; (i) DSIFN; (j) DESSN; (k)BIT; (l)VcT; (m)DMINet; (n)GeSANet; (o)EATDer; (p)TCD-Net; (q)LRNet. For convenience, several colors are used to facilitate a clearer visualization of results: i.e., TP (white), TN (black), \textcolor{green}{FP (green)} and \textcolor[RGB]{255,0,255}{FN (purple)}.}
	\label{fig8}
\end{figure*}

\begin{figure*}[!t]
	\centering
	\includegraphics[width=6.8in]{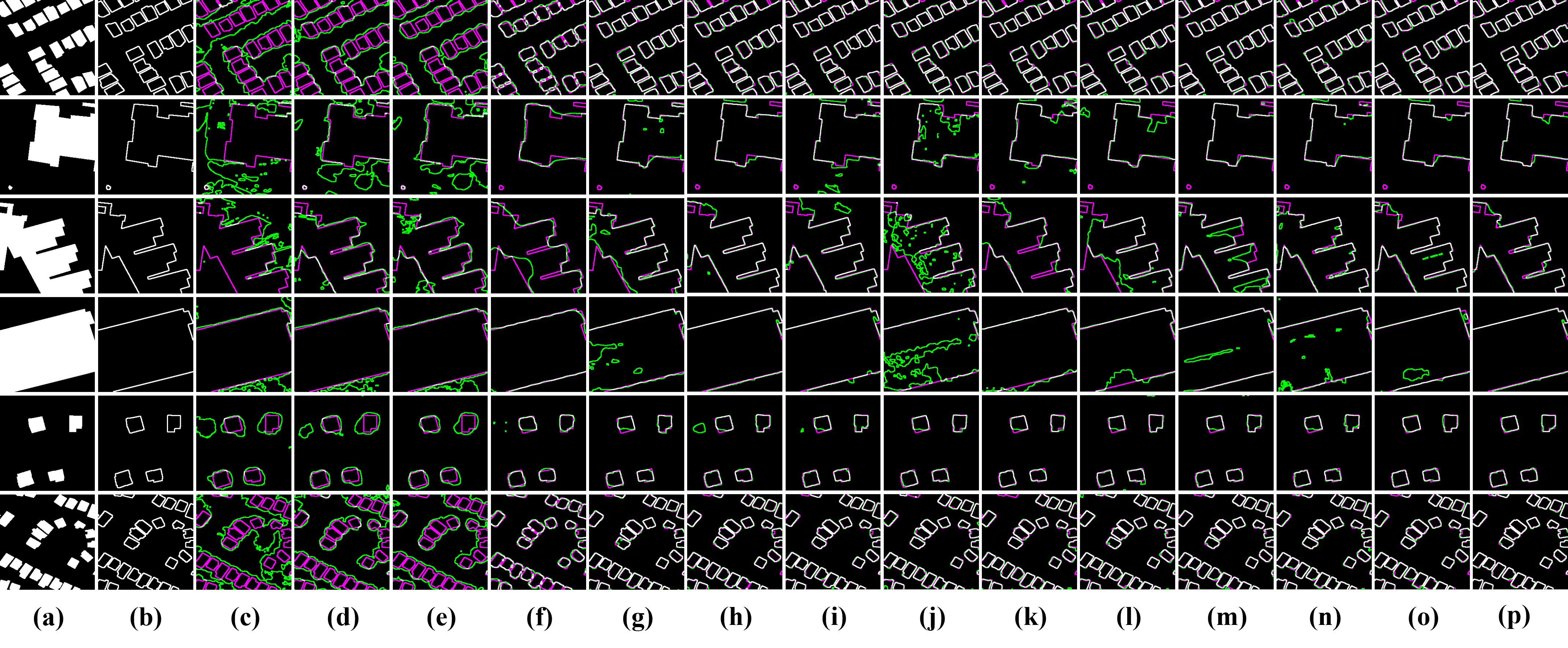}
	\caption{Visualization of edges in the change detection results on the LEVIR-CD dataset (a) Label area; (b) Label edge; (c) FC-EF; (d) FC-Siam-Conc; (e) FC-Siam-Diff; (f) DASNet; (g) SNUNet; (h) DSIFN; (i) DESSN; (j)BIT; (k)VcT; (l)DMINet; (m)GeSANet; (n)EATDer; (o)TCD-Net; (p)LRNet. For convenience, several colors are used to facilitate a clearer visualization of results: i.e., TP (white), TN (black), \textcolor{green}{FP (green)} and \textcolor[RGB]{255,0,255}{FN (purple)}.}
	\label{fig9}
\end{figure*}

To further evaluate the change detection performance of different models on the LEVIR-CD dataset, the test results are visualized and compared with the original bi-temporal images as well as the labels, as shown in Fig. \ref{fig8} (area) and Fig. \ref{fig9} (edge). From Fig. \ref{fig8}, it can be noticed that the three baseline models (d-f) exhibit numerous false detections, indicating that their simplistic design struggles to analyze complex features. BiT (k), as a ViT-based model, focuses more on global information, which leads to an imbalance of attention with the local information, resulting in a large number of missed detections. Other attention-based or ViT-based models introduce different modules to improve the identification and attention to change areas, significantly reducing false and missed detections of changed objects. The proposed LRNet, employing a localization-then-refinement strategy, achieves accurate localization of change areas and effectively improves edge detection accuracy, thereby enhancing overall performance. The visualization results of LRNet are the closest to those of labels.

Fig. \ref{fig9} illustrates the edge discrimination performance of different models. It can be found that the compared models have more or less some bias in the discrimination of edges. Models c-e barely detect any edges. Models f-o detect most edges but still exhibit significant deviations in the interior and details of change areas. The proposed LRNet accomplishes the identification and detection of change edges well through accurate localization of change areas by the change alignment modules as well as the depth-to-light refinement. Corresponding to the quantitative results in Table \ref{tab2}, the discrimination results of change edges by LRNet match the labels the most.

2) Comparison experiments on the WHU-CD dataset: Table \ref{tab3} presents the quantitative results of each model on the WHU-CD dataset. Compared to the other compared methods, LRNet achieves the best results in the evaluation metrics of change areas: OA (99.47\%), Pre (95.11\%), F1 (92.51\%), and IOU (86.06\%). FC-Siam-Conc exhibits the highest Rec (95.45\%), but its Pre (45.54\%) is significantly lower than other SOTA methods and LRNet. Considering both Pre and Rec, the proposed LRNet achieves the best F1, with improvements ranging from 0.95\% to 40.56\% compared to other models. Only LRNet and TCD-Net achieve F1 values above 90\% among all models, indicating that the three-branch structure can effectively extract and retain key information from original bi-temporal images and their difference features. For the discrimination of change edges, similar to the results on the LEVIR-CD dataset, LRNet achieves the best values in all four evaluation metrics: $Pre_{Edge}$ (68.01\%), $Rec_{Edge}$ (63.56\%), $F1_{Edge}$ (65.71\%), and $IOU_{Edge}$ (48.93\%), with improvements of 10.24\%, 3.59\%, 9.15\%, and 9.50\%, respectively, compared to the suboptimal values. The significant improvement in quantitative evaluation metrics demonstrates the excellent change detection performance of LRNet.

\begin{table*}[!t]
	\centering
	\caption{Evaluation metrics of each change detection model on the WHU-CD dataset.}
	\label{tab3}
	
	\begin{tabular}{cccccccccc}
		\toprule
		Model & OA(\%) & Pre(\%) & Rec(\%) & F1(\%) & IOU(\%) & $Pre_{Edge}$(\%) & $Rec_{Edge}$(\%) & $F1_{Edge}$(\%) & $IOU_{Edge}$(\%) \\
		\midrule
		FC-EF & 93.74 & 35.95 & 93.60 & 51.95 & 35.09 & 2.86 & 22.09 & 5.07 & 2.60 \\ 
		FC-Siam-Conc & 95.70 & 45.54 & \textbf{95.45} & 61.66 & 44.58 & 3.48 & 14.33 & 5.60 & 2.88 \\ 
		FC-Siam-Diff & 97.07 & 55.85 & 91.75 & 69.43 & 53.18 & 8.25 & 23.30 & 12.18 & 6.49 \\ 
		DASNet & 99.11 & 86.74 & 89.23 & 87.97 & 78.53 & 6.88 & 12.91 & 4.70 & 8.98 \\ 
		SNUNet & 99.23 & 89.48 & 89.28 & 89.38 & 80.81 & 41.21 & 57.52 & 48.02 & 31.59 \\ 
		DSIFN & 99.15 & 86.51 & 90.73 & 88.57 & 79.49 & 47.48 & 48.66 & 48.06 & 31.63 \\ 
		DESSN & 99.19 & 89.60 & 87.98 & 88.78 & 79.84 & 43.05 & 59.97 & 50.12 & 33.44 \\ 
		BIT & 99.20 & 90.21 & 87.35 & 88.76 & 79.78 & 42.01 & 54.29 & 47.37 & 31.04 \\ 
		VcT & 99.06 & 86.98 & 87.21 & 87.09 & 77.14 & 35.62 & 43.91 & 39.34 & 24.48 \\ 
		DMINet & 99.27 & 92.96 & 86.38 & 89.55 & 81.08 & 53.65 & 55.12 & 54.37 & 37.34 \\ 
		GeSANet & 99.26 & 90.51 & 89.06 & 89.78 & 81.46 & 52.34 & 52.10 & 52.22 & 35.34 \\ 
		EATDer & 99.23 & 90.70 & 87.54 & 89.09 & 80.33 & 52.16 & 59.49 & 55.58 & 38.49 \\ 
		TCD-Net & 99.39 & 92.02 & 91.12 & 91.56 & 84.45 & 57.77 & 55.39 & 56.56 & 39.43 \\ 
		LRNet & \textbf{99.47} & \textbf{95.11} & 90.04 & \textbf{92.51} & \textbf{86.06} & \textbf{68.01} & \textbf{63.56} & \textbf{65.71} & \textbf{48.93} \\ 
		\bottomrule
	\end{tabular}
	
\end{table*}

\begin{figure*}[!t]
	\centering
	\includegraphics[width=6.8in]{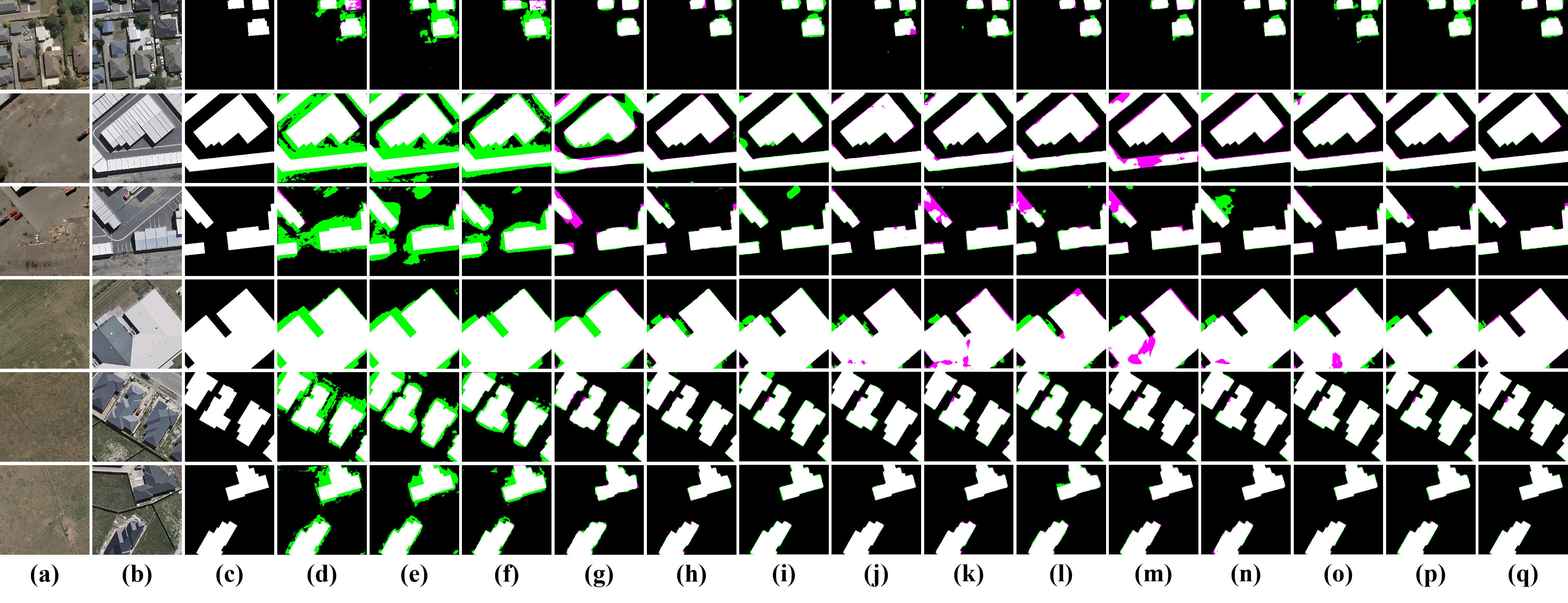}
	\caption{Visualization of change detection results on the WHU-CD dataset (a) T1 image; (b) T2 image; (c) Label; (d) FC-EF; (e) FC-Siam-Conc; (f) FC-Siam-Diff; (g) DASNet; (h) SNUNet; (i) DSIFN; (j) DESSN; (k)BIT; (l)VcT; (m)DMINet; (n)GeSANet; (o)EATDer; (p)TCD-Net; (q)LRNet. For convenience, several colors are used to facilitate a clearer visualization of results: i.e., TP (white), TN (black), \textcolor{green}{FP (green)} and \textcolor[RGB]{255,0,255}{FN (purple)}.}
	\label{fig10}
\end{figure*} 

\begin{figure*}[!t]
	\centering
	\includegraphics[width=6.8in]{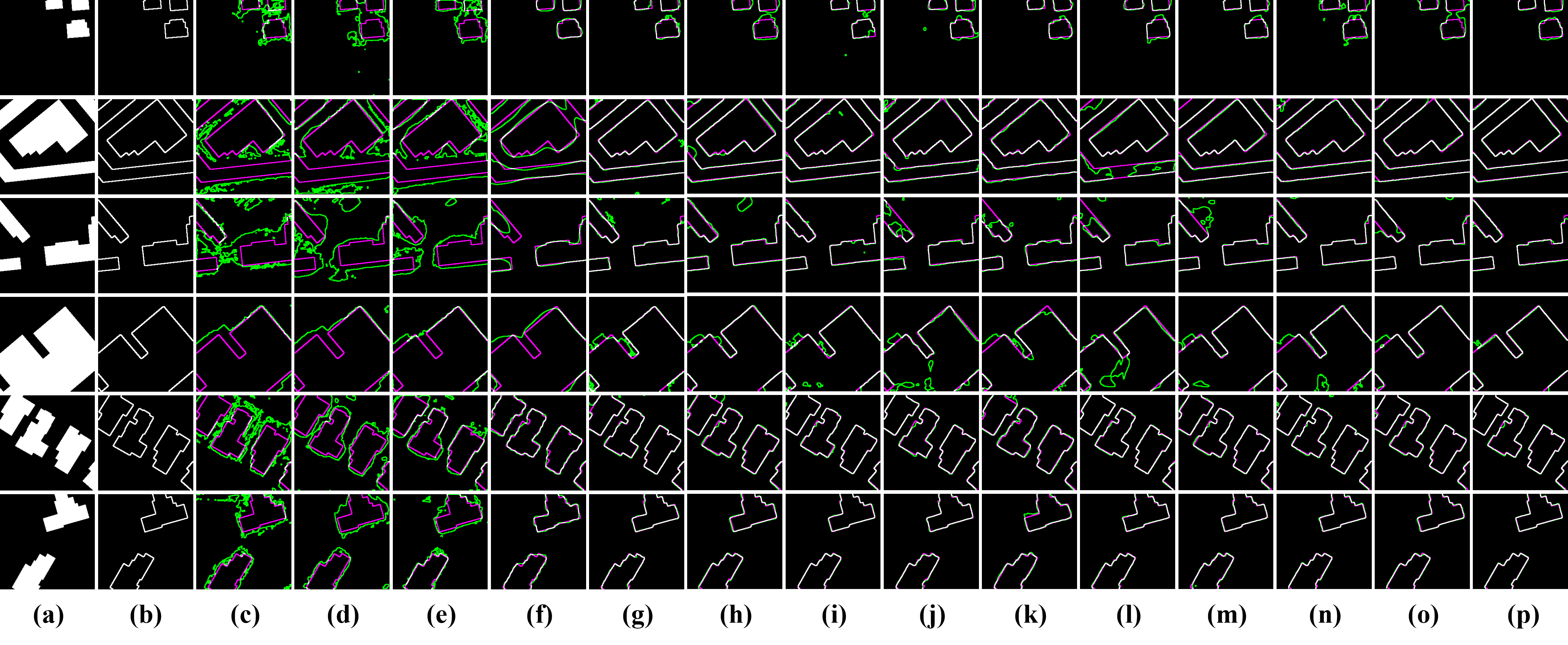}
	\caption{Visualization of edges in the change detection results on the WHU-CD dataset (a) Label area; (b) Label edge; (c) FC-EF; (d) FC-Siam-Conc; (e) FC-Siam-Diff; (f) DASNet; (g) SNUNet; (h) DSIFN; (i) DESSN; (j)BIT; (k)VcT; (l)DMINet; (m)GeSANet; (n)EATDer; (o)TCD-Net; (p)LRNet. For convenience, several colors are used to facilitate a clearer visualization of results: i.e., TP (white), TN (black), \textcolor{green}{FP (green)} and \textcolor[RGB]{255,0,255}{FN (purple)}.}
	\label{fig11}
\end{figure*} 

To further evaluate the change detection performance of different models on the WHU-CD dataset, the test results are visualized and compared with the original bi-temporal images as well as the labels, as shown in Fig. \ref{fig10} (area) and Fig. \ref{fig11} (edge). From Fig. \ref{fig10}, it can be seen that all methods can accurately locate the positions of change areas, but there are significant discrepancies in the discrimination of edge details. Models d-f have difficulty in discriminating pseudo-changes and uninteresting changes, leading to numerous false detections. Since the WHU-CD dataset focuses on changes in buildings, models d-f incorrectly detect roads or other types of changes. Other attention-based or ViT-based methods effectively differentiate between different types of changes by enhancing attention to change areas and extracting global information, thus reducing the number of false detections. The proposed LRNet interacts change features extracted from the three branches and utilizes change alignment modules to align change objects with constraints and accurately locate them. And then the change edges are refined from deep to shallow, which effectively balances the number of false detections and missed detections to achieve the optimal results. Fig. \ref{fig11} illustrates the edge discrimination performance of different models. It can be observed that compared models more or less have some bias in edge discrimination, which is manifested as false detection (green) and missed detection (purple). The edge discrimination results of the proposed LRNet are almost entirely white, i.e., they are identical to the labels, with only a little deviation in very few cases, significantly outperforming other methods. These quantitative and qualitative results demonstrate that the proposed LRNet effectively utilizes the features extracted by the three-branch encoder for the localization and refinement of changes, significantly improving the accuracy of change detection model, particularly in the discrimination of change edges.

\subsection{Ablation study}

In this section, the effectiveness of key components for LRNet will be discussed, including LOP, C2A, HCA, and E2A. Since HCA works on top of C2A, C2A and HCA are discussed together as a whole. The baseline, "Model base", is a three-branch change detection model with VGG16 as the backbone. Ablation experiments are conducted on both datasets. As OA, F1, and IOU can evaluate the overall performance of the model, they are selected as evaluation metrics for the ablation experiments, along with $F1_{Edge}$ and $IOU_{Edge}$. The results of ablation experiments are presented in Table \ref{tab4}.

\begin{table*}
	\centering
	\caption{Results of ablation experiments.}
	\label{tab4}
	\resizebox{\textwidth}{!}{

	\begin{tabular}{ccccccccccccccc}
		\toprule
		\multirow{2}{*}{Model} & \multirow{2}{*}{LOP} & \multirow{2}{*}{C2A} & \multirow{2}{*}{HCA} & \multirow{2}{*}{E2A}  & \multicolumn{5}{c}{LEVIR-CD} & \multicolumn{5}{c}{WHU-CD} \\
		\cmidrule{6-15}
		 &  &  &  &  & OA(\%) & F1(\%) & IOU(\%) & $F1_{Edge}$(\%) & $IOU_{Edge}$(\%) & OA(\%) & F1(\%) & IOU(\%) & $F1_{Edge}$(\%) & $IOU_{Edge}$(\%) \\ 
		\midrule
		Model base &  &  &  &  & 98.99 & 89.96 & 81.75 & 67.03 & 51.71 & 99.35 & 90.60 & 82.81 & 62.94 & 45.92 \\ 
		model1 (base+ ) & \checkmark &  &  &  & 99.01 & 90.44 & 82.55 & 68.03 & 51.55 & 99.39 & 91.39 & 84.16 & 63.85 & 46.98 \\ 
		model2 (base+ ) &  & \checkmark &  &  & 99.04 & 90.27 & 82.27 & 69.04 & 52.72 & 99.38 & 91.24 & 83.89 & 63.17 & 46.17 \\ 
		model3 (base+ ) &  & \checkmark & \checkmark &  & 99.04 & 90.62 & 82.85 & 68.44 & 52.02 & 99.41 & 91.69 & 84.66 & 63.68 & 46.52 \\ 
		model4 (base+ ) &  &  &  & \checkmark & 99.06 & 90.54 & 82.71 & 69.54 & 53.49 & 99.40 & 91.48 & 84.30 & 64.86 & 47.84 \\ 
		model5 (base+ ) & \checkmark & \checkmark & \checkmark &  & 99.07 & 90.75 & 83.06 & 69.26 & 52.97 & 99.44 & 92.15 & 85.43 & 63.84 & 46.72 \\ 
		model6 (base+ ) & \checkmark &  &  & \checkmark & 99.08 & 90.95 & 83.42 & 70.31 & 54.22 & 99.43 & 91.87 & 84.96 & 64.91 & 48.03 \\ 
		model7 (base+ ) &  & \checkmark & \checkmark & \checkmark & 99.07 & 90.76 & 83.08 & 69.91 & 53.69 & 99.45 & 92.29 & 85.68 & 65.16 & 48.26 \\ 
		LRNet & \checkmark & \checkmark & \checkmark & \checkmark & \textbf{99.10} & \textbf{91.08} & \textbf{83.63} & \textbf{70.95} & \textbf{54.78} & \textbf{99.47} & \textbf{92.51} & \textbf{86.06} & \textbf{65.71} & \textbf{48.93} \\ 
		\bottomrule
	\end{tabular}

	}
	
\end{table*}

1) Effectiveness of LOP: To reduce information loss during feature extraction, LOP is proposed. As shown in Table \ref{tab4}, Model1, with the addition of LOP to the Model base, exhibits improvements in all four metrics (OA, F1, IOU, and $F1_{Edge}$) on both datasets. Model5, built upon Model3 with the LOP module added, demonstrates significant improvements in all metrics on both datasets. Similarly, Model6 and Model4, LRNet and Model7 also show the same trend. These findings suggest that the inclusion of LOP effectively reduces the loss of critical information, which in turn assists the model in identifying and detecting changes, thus improving the accuracy of change detection and validating the effectiveness of LOP.

2) Effectiveness of C2A and HCA: For the interaction of change features from different branches and the fusion of features from different levels to identify targets of different sizes, C2A and HCA are designed. Model2 adds the C2A module to the Model base as an assistant to the interaction and fusion of change features. All the metrics on both datasets are significantly improved, indicating that C2A effectively interacts the features from different branches, aligning and strengthening the change areas. Compared to Model2, Model3 has an additional HCA module, which propagates the change alignment attention map to different levels in a hierarchical and densely connected manner for precise localization of changes of different sizes. Model3 exhibits improvements in area metrics (OA, F1, IOU) on both datasets, while the edge metrics ($F1_{Edge}$ and $IOU_{Edge}$) decrease on the LEVIR-CD dataset. This suggests that the addition of HCA helps the localization and identification of change areas in different scenarios but may not be favorable for edge discrimination when the edge is unconstrained.

Considering C2A and HCA together, Model3 has a significant improvement in all metrics compared to the Model base with the additional combination of C2A and HCA. Similar trends also occur between Model5 and Model1, Model7 and Model4, LRNet and Model6. These results indicate that the combination of C2A and HCA can well perform the interaction and fusion of change features, assisting in the localization and identification of change areas, which in turn improves the detection performance of the model. Additionally, this confirms the effectiveness of the proposed C2A and HCA.

3) Effectiveness of E2A: To constrain and correct change areas and edges in the localization results, the E2A module is proposed. Similarly, we constrain the inference results of LRNet in the same manner. Model4, Model6, Model7 and LRNet are obtained by adding the E2A module to the Model base, Model1, Model3 and Model5, respectively. As shown in Table \ref{tab4}, all the evaluation metrics on both datasets of the models with an additional E2A module are increased compared to the corresponding models without the addition. Notably, the edge metrics, $F1_{Edge}$ and $IOU_{Edge}$, show significant improvements, with gains of 2.51\% and 2.67\%, respectively. These results reflect the effective constraint of the E2A module on the change areas and edges, validating its effectiveness.

Sequentially adding the proposed modules to the Model base, it can be noticed that the overall metrics on both datasets are significantly improved. Particularly, the E2A module plays a crucial role in edge discrimination, improving the detection accuracy of the areas when incorporated, with a larger increase in edge discrimination accuracy compared to that of the areas. The best performance is the base+LOP+C2A+HCA+E2A combination, i.e., the proposed network LRNet. The above results and findings well validate the outstanding change detection performance of LRNet, as well as the effectiveness and necessity of each module.

\subsection{Discussion}
\label{discussion_4_6}
1) Discussion on Loss Functions: In this section, we evaluate the difference in the effectiveness of the models trained using BCE loss, IOU loss, and their joint loss, respectively. From Table \ref{tab5}, it can be noticed that the proposed model achieves the best performance when trained using the joint BCE and IOU loss. The model performs worst when trained using BCE loss alone. Experimental results indicate that the joint loss can improve the accuracy of the model taking advantage of the constraints from different aspects of the BCE and IOU, where the IOU takes more into account the segmentation accuracy as well as the focus on fewer category for change areas. Additionally, it is noteworthy that the use of joint loss improves the accuracy of change edge detection significantly more than that of change area detection.
\begin{table*}
	\centering
	\caption{Discussion results of loss functions.}
	\label{tab5}
	\resizebox{\textwidth}{!}{
		\begin{tabular}{ccccccccccc}
			\toprule
			\multirow{2}{*}{Loss} & \multicolumn{5}{c}{LEVIR-CD} & \multicolumn{5}{c}{WHU-CD} \\
			\cmidrule{2-11}
			& OA(\%) & F1(\%) & IOU(\%) & $F1_{Edge}$(\%) & $IOU_{Edge}$(\%) & OA(\%) & F1(\%) & IOU(\%) & $F1_{Edge}$(\%) & $IOU_{Edge}$(\%) \\ 
			\midrule
			BCE & 98.89 & 88.47 & 79.32 & 62.84 & 45.82 & 99.39 & 91.33 & 84.05 & 59.58 & 42.43 \\ 
			IOU & 98.97 & 89.83 & 81.54 & 66.72 & 50.06 & 99.45 & 92.28 & 85.68 & 62.85 & 45.71 \\ 
			BCE+IOU & \textbf{99.10} & \textbf{91.08} & \textbf{83.63} & \textbf{70.95} & \textbf{54.78} & \textbf{99.47} & \textbf{92.51} & \textbf{86.06} & \textbf{65.71} & \textbf{48.93} \\  
			\bottomrule
		\end{tabular}
	}
\end{table*}

2) Complexity Analysis: For a comprehensive and fair comparison of the performance, we evaluate the complexity of the models using the number of parameters (Params.) and floating-point operations (FLOPs). Furthermore, the ratio of FLOPs to Params. is introduced to measure the efficiency of the parameters, where a lower value indicates that the increase in parameters brings about  a proportionally smaller increase in computational load. Table \ref{tab6} presents the complexity metrics of different models along with their comprehensive evaluation results on each dataset. It can be found that FC-EF, FC-Siam-Conc, and FC-Siam-Diff have the lowest FLOPs and Params., but they also exhibit the poorest performance on both datasets. Other compared models correspondingly improve the feature extraction capability and increase the model complexity with the introduction of deeper network structures, attention mechanisms, deep supervision, or ViT modules. These models show an increase in both FLOPs and Params., and simultaneously achieve significantly improved change detection accuracy. Although the proposed LRNet utilizes three branches for feature extraction, resulting in the highest FLOPs and Params., its ratio of FLOPs to Params. is the second lowest, indicating that the parameter increase of LRNet is efficient and does not proportionally increase computational load. Notably, while maintaining a reasonably acceptable Params. and FLOPs, the proposed LRNet achieves the best comprehensive evaluation results on both datasets.
\begin{table*}
	\centering
	\caption{Model complexity metrics versus comprehensive evaluation metrics on each dataset.}
	\label{tab6}
	\resizebox{\textwidth}{!}{
		\begin{tabular}{cccccccccccccc}
			\toprule
			\multirow{2}{*}{Model} & \multirow{2}{*}{FLOPs(G)} & \multirow{2}{*}{Params.(K)} & \multirow{2}{*}{FLOPs/Params.(K)} & \multicolumn{5}{c}{LEVIR-CD} & \multicolumn{5}{c}{WHU-CD} \\
			\cmidrule{5-14}
			&  &  &  & OA(\%) & F1(\%) & IOU(\%) & $F1_{Edge}$(\%) & $IOU_{Edge}$(\%) & OA(\%) & F1(\%) & IOU(\%) & $F1_{Edge}$(\%) & $IOU_{Edge}$(\%) \\ 
			\midrule
			FC-EF & \textbf{3.13} & \textbf{1.35} & 2.32 & 90.72 & 51.28 & 34.49 & 6.63 & 3.43 & 93.74 & 51.95 & 35.09 & 5.07 & 2.60 \\ 
			FC-Siam-Conc & 4.89 & 1.55 & 3.15 & 94.61 & 64.59 & 47.71 & 10.14 & 5.34 & 95.7 & 61.66 & 44.58 & 5.60 & 2.88 \\ 
			FC-Siam-Diff & 4.29 & 1.35 & 3.18 & 96.06 & 70.41 & 54.34 & 12.51 & 6.67 & 97.07 & 69.43 & 53.18 & 12.18 & 6.49 \\ 
			DASNet & 56.92 & 16.26 & 3.50 & 98.43 & 85.09 & 74.05 & 15.48 & 8.39 & 99.11 & 87.97 & 78.53 & 4.70 & 8.98 \\ 
			SNUNet & 11.79 & 3.01 & 3.92 & 98.92 & 89.22 & 80.54 & 66.46 & 49.77 & 99.23 & 89.38 & 80.81 & 48.02 & 31.59 \\ 
			DSIFN & 79.04 & 35.99 & 2.20 & 98.95 & 89.48 & 80.97 & 64.95 & 48.1 & 99.15 & 88.57 & 79.49 & 48.06 & 31.63 \\ 
			DESSN & 27.10 & 18.35 & 1.48 & 99.01 & 90.06 & 81.92 & 68.10 & 51.63 & 99.19 & 88.78 & 79.84 & 50.12 & 33.44 \\ 
			BIT & 10.63 & 3.50 & 3.04 & 98.84 & 88.3 & 79.04 & 63.32 & 46.33 & 99.2 & 88.76 & 79.78 & 47.37 & 31.04 \\ 
			VcT & 10.64 & 3.57 & 2.98 & 98.90 & 89.18 & 80.47 & 63.89 & 46.94 & 99.06 & 87.09 & 77.14 & 39.34 & 24.48 \\ 
			DMINet & 14.55 & 6.75 & 2.16 & 98.94 & 89.42 & 80.86 & 66.55 & 49.87 & 99.27 & 89.55 & 81.08 & 54.37 & 37.34 \\ 
			GeSANet & 72.57 & 36.50 & 1.99 & 99.03 & 90.32 & 82.35 & 67.87 & 51.36 & 99.26 & 89.78 & 81.46 & 52.22 & 35.34 \\ 
			EATDer & 23.43 & 6.59 & 3.56 & 98.94 & 89.59 & 81.15 & 66.17 & 49.44 & 99.23 & 89.09 & 80.33 & 55.58 & 38.49 \\ 
			TCD-Net & 56.64 & 11.13 & \textbf{5.09} & 99.04 & 90.37 & 82.43 & 68.50 & 52.09 & 99.39 & 91.56 & 84.45 & 56.56 & 39.43 \\ 
			LRNet & 92.23 & 48.71 & 1.89 & \textbf{99.10} & \textbf{91.08} & \textbf{83.63} & \textbf{70.95} & \textbf{54.78} & \textbf{99.47} & \textbf{92.51} & \textbf{86.06} & \textbf{65.71} & \textbf{48.93} \\  
			\bottomrule
		\end{tabular}
	}
\end{table*}

\section{Conclusion}
\label{Conclusion}
In this study, a novel change detection network based on a localization-then-refinement strategy, namely LRNet, is proposed to address the challenge of discriminating change edges. LRNet consists of two stages: localization and refinement. In the localization stage, a three-branch encoder extracts features from the original images and their differences from shallow to deep, progressively locating the position of each changed area. LOP is proposed to reduce information loss during feature extraction and participate in the optimization of the entire network. The C2A and HCA modules are proposed to interact the change feature information from different branches and accurately locate different sizes of changed areas. In the refinement stage, E2A is designed to constrain and refine the localization results. By combining the difference features enhanced by C2A from deep to shallow, different sizes of changed areas and their edges are refined, ultimately obtaining accurate results. The proposed LRNet achieves the best comprehensive evaluation metrics and the most precise edge discrimination results compared with other 13 state-of-the-art methods on the LEVIR-CD and WHU-CD datasets. However, the current model still has some shortcomings. While achieving optimal detection accuracy, the overall complexity of the proposed LRNet is relatively high. In the future, we will explore lightweight backbones or knowledge distillation methods to reduce the complexity while maintaining high detection accuracy. Additionally, we will investigate more advanced and efficient techniques to support more complex scenarios.

\section*{Declaration of Competing Interest}
The authors declare that they have no known competing financial interests or personal relationships that could have appeared to influence the work reported in this paper. 

\section*{Acknowledgment}
This work was supported by the National Natural Science Foundation of China (No. T2122014), and the National Key Research and Development Program of China (No. 2022YFB3903300). The numerical calculations in this paper have been done on the supercomputing system in the Supercomputing Center of Wuhan University.

\bibliographystyle{cas-model2-names}
%
\bibliography{refs.bib}

%
%

\end{document}